\documentclass[11pt, twocolumn]{article}
\usepackage[utf8]{inputenc}
\usepackage{lmodern}
\usepackage{geometry}
\usepackage{amsmath, amssymb}
\usepackage{graphicx}
\usepackage{url}
\usepackage{authblk}
\usepackage{hyperref}
\usepackage{algorithm}
\usepackage{algpseudocode}
\usepackage{caption}
\usepackage{float}
\usepackage{cleveref}
\usepackage{placeins}
\usepackage{xcolor}
\usepackage[authoryear,round]{natbib}
\usepackage{framed,multirow}

\crefname{table}{Tab.}{Tables}
\crefname{figure}{Fig.}{Figs.}

\hypersetup{
    colorlinks=true,
    linkcolor=blue,
    citecolor=blue,
    urlcolor=blue
}
\usepackage{dsfont}
\usepackage{xspace}
\usepackage{booktabs}
\usepackage{pifont}
\usepackage{gensymb}
\usepackage{hhline}
\usepackage{makecell}
\usepackage{tabularx}
\usepackage{lipsum}

\newcommand{\cmark}{\ding{51}} %
\newcommand{\xmark}{\ding{55}} %

\usepackage[usenames,dvipsnames]{xcolor}

\newcommand{\blockcomment}[1]{}

\makeatletter
\newcommand{\FIXME}{\@ifstar{\FIXME@star}{\FIXME@nostar}}
\newcommand{\FIXME@nostar}[1]{\colorbox{red}{\textbf{FIXME:}~#1}\xspace} 
\newcommand{\FIXME@star}[1]{\colorbox{red}{#1}\xspace}

\newcommand{\TODO}{\@ifstar{\TODO@star}{\TODO@nostar}}
\newcommand{\TODO@nostar}[1]{\colorbox{yellow}{\textbf{TODO:}~#1}\xspace} 
\newcommand{\TODO@star}[1]{\colorbox{yellow}{#1}\xspace}
\newcommand{\UPD}[1]{#1\xspace}

\makeatother

\newcommand{\ours}{OCELOT\xspace}

\newcommand{\saltfish}{\cite{li2023enhancing}\xspace}
\newcommand{\joshua}{\cite{millward2023dense}\xspace}
\newcommand{\lysch}{\cite{schoenpflug2023softctm}\xspace}
\newcommand{\biototem}{\cite{Zheng2023enhanced}\xspace}
\newcommand{\yllab}{\cite{lo2023enhancing}\xspace}
\makeatother

\title{OCELOT 2023: Cell Detection from Cell-Tissue Interaction Challenge\thanks{OCELOT stands for \textit{\textbf{O}verlapped \textbf{Cel}l \textbf{o}n \textbf{T}issue Dataset}.}}
\author[1]{JaeWoong Shin}
\author[1]{Jeongun Ryu}
\author[1]{Aaron Valero Puche}
\author[1]{Jinhee Lee}
\author[1]{Biagio Brattoli}
\author[1]{Wonkyung Jung}
\author[1]{Soo Ick Cho}
\author[1]{Kyunghyun Paeng}
\author[1]{Chan-Young Ock}
\author[1]{Donggeun Yoo}
\author[2]{Zhaoyang Li}
\author[2]{Wangkai Li}
\author[2]{Huayu Mai}
\author[3]{Joshua Millward}
\author[3]{Zhen He}
\author[3]{Aiden Nibali}
\author[4]{Lydia Anette Schoenpflug}
\author[4,5,6]{Viktor Hendrik Koelzer}
\author[7]{Xu Shuoyu}
\author[7]{Ji Zheng}
\author[7]{Hu Bin}
\author[8]{Yu-Wen Lo}
\author[8]{Ching-Hui Yang}
\author[1]{Sérgio Pereira\thanks{Corresponding author: \href{mailto:sergio@lunit.io}{sergio@lunit.io}}}

\affil[1]{Lunit Inc., Seoul, Republic of Korea}
\affil[2]{University of Science and Technology of China, Hefei, China}
\affil[3]{School of Computing, Engineering and Mathematical Sciences, La Trobe University, Melbourne, Australia}
\affil[4]{Department of Pathology and Molecular Pathology, University Hospital of Z\"{u}rich, University of Z\"{u}rich, Z\"{u}rich, Switzerland}
\affil[5]{Institute of Medical Genetics and Pathology, University Hospital Basel, University of Basel, Basel, Switzerland}
\affil[6]{Department of Oncology, University of Oxford, Oxford, UK}
\affil[7]{Bio-totem Pte Ltd, Foshan, China}
\affil[8]{Department of Computer Science, National Tsing Hua University, Hsinchu, Taiwan}
\date{}

\begin{document}

\twocolumn[
{\noindent\small\color{gray}
\begin{center}  
This is the accepted manuscript of an article published in \textit{Medical Image Analysis} (Elsevier).\\
The final version is available at: \url{https://doi.org/10.1016/j.media.2025.103751}. \\
\copyright\ 2025. CC BY-NC-ND 4.0 license.
\end{center}
}
\vspace{-2em}
\maketitle
\vspace{-2em}
\begin{abstract}
Pathologists routinely alternate between different magnifications when examining Whole-Slide Images, allowing them to evaluate both broad tissue morphology and intricate cellular details to form comprehensive diagnoses. However, existing deep learning-based cell detection models struggle to replicate these behaviors and learn the interdependent semantics between structures at different magnifications. A key barrier in the field is the lack of datasets with multi-scale overlapping cell and tissue annotations. The OCELOT 2023 challenge was initiated to gather insights from the community to validate the hypothesis that understanding cell and tissue (cell-tissue) interactions is crucial for achieving human-level performance, and to accelerate the research in this field. The challenge dataset includes overlapping cell detection and tissue segmentation annotations from six organs, comprising 673 pairs sourced from 306 The Cancer Genome Atlas (TCGA) Whole-Slide Images with hematoxylin and eosin staining, divided into training, validation, and test subsets. Participants presented models that significantly enhanced the understanding of cell-tissue relationships. Top entries achieved up to a 7.99 increase in F1-score on the test set compared to the baseline cell-only model that did not incorporate cell-tissue relationships. This is a substantial improvement in performance over traditional cell-only detection methods, demonstrating the need for incorporating multi-scale semantics into the models. This paper provides a comparative analysis of the methods used by participants, highlighting innovative strategies implemented in the OCELOT 2023 challenge.
\end{abstract}

\vspace{4em}
]

\section{Introduction}

Deep learning has revolutionized the field of computational pathology, offering unprecedented capabilities in the analysis and interpretation of Whole Slide Images (WSI). These advancements allow more accurate and efficient identification of disease-related \citep{komura2018machine} features at both cellular \citep{coudray2018classification,sirinukunwattana2016locality} and tissue levels \citep{gelasca2008evaluation,wang2016deep}. Hence, it unlocks the potential for advances in the understanding and treatment of cancer, such as finding and developing new biomarkers \citep{mobadersany2018predicting,skrede2020deep}, streamlining the identification of cancer driver mutations \citep{schaumberg2016h}, tumor purity quantification \citep{gong2020tumor}, or automatic cancer grading \citep{bulten2020automated}. These technologies, ultimately, hold the potential for personalized medicine and better patient outcomes.

 In computational pathology, two critical quantification tasks are cell detection and tissue segmentation. 
Cell detection involves localizing and classifying various cell types within histology images, providing essential information for understanding cellular composition and distribution \citep{sirinukunwattana2016locality} and serving as valuable quantitative biomarkers \citep{lara2021quantitative}.
Accurate tumor cell detection has demonstrated significant clinical value in recent applications, including tumor purity assessment for molecular analysis in colorectal cancer \citep{schoenpflug2025tumour} and AI-powered HER2 scoring in breast cancer screening \citep{kapil2024her2}.
Additionally, lymphocyte detection can be extended to tumor-infiltrating lymphocyte evaluation and therapeutic response prediction~\citep{choi2023deep}.

Tissue segmentation, on the other hand, involves delineating different tissue regions, enabling the analysis and identification of tissue architectures \citep{ronneberger2015u}. By combining these tasks, researchers can develop composite biomarkers that capture the relationships between cellular and larger tissue-level features, offering deeper insights into disease mechanisms and potential therapeutic targets \citep{ma2017data, park2022artificial}.

Traditional cell detection models, however, face significant challenges. Because of the large size of WSIs and the need for assessing fine details of cells, these models typically operate on high-magnification images with a limited Field-of-View (FoV), which restricts their ability to contextualize cell types within broader tissue structures. This narrow focus can lead to an over-reliance on appearance details without considering the spatial arrangement and interactions of cells within the tissue area. In contrast, pathologists overcome this limitation by dynamically zooming in and out during their examinations, thus integrating both macro and micro perspectives to form holistic assessments. This multi-scale approach enables them to understand cellular features within the context of the surrounding tissue. However, such capability has seldom been addressed within deep learning models. While some methods have attempted to use a larger FoV \citep{van2021hooknet,bai2020multi,yuan2024new}, these methods lack the necessary labels to enforce semantic understanding, which is important for accurate cell-tissue relationships interpretation as suggested by \cite{ryu2023ocelot}.

To address the lack of public datasets designed specifically to leverage cell-tissue relationships for cell detection and accelerate this research field, the ``OCELOT 2023: Cell Detection from Cell-Tissue Interaction'' (OCELOT 2023) challenge was organized. This challenge aimed to gather community insights and validate the hypothesis that understanding cell-tissue interactions is crucial for achieving superior performance in computational pathology tasks. The challenge dataset, sourced from 306 TCGA~\citep{HUTTER2018283} WSIs with hematoxylin and eosin (H\&E) staining across six organs, included overlapping cell detection and tissue segmentation annotations. Participants were tasked with developing models that leverage these multi-scale annotations to enhance the understanding of cell-tissue relationships. Proposals leveraging cell-tissue relationships significantly improved model performance. Top entries achieved substantial increases in F1-scores, up to 7.99 points higher than the cell-only baseline that did not utilize these relationships. This outcome confirms the hypothesis that incorporating the relationships between cells and tissues can lead to more accurate and effective models. This paper presents a comparative analysis of the methods used by participants, highlighting the innovative strategies that emerged from the OCELOT 2023 challenge.

The paper is organized as follows. In Section \ref{sec:ocelot}, we introduce the OCELOT 2023 challenge, including its goals, dataset, and organization of the challenge. In Section \ref{sec:submissions} we describe the top-performing submissions and their key technical components. In Section \ref{sec:results} we present the results of the challenge. Finally, in Section \ref{sec:conclusion} the conclusions and future directions are outlined.
The paper adheres to the Biomedical Image Analysis ChallengeS (BIAS) guideline \citep{maier2020bias}, providing transparent and standardized reporting of our challenge design, execution, and results.

\section{OCELOT 2023 Challenge}
\label{sec:ocelot}

\begin{table}
\small

\centering
\caption{Number of slides and cell-tissue patch pairs per organ in the OCELOT dataset. \UPD{$\dagger$: Kidney specimens comprised the following histological subtypes: chromophobe renal cell carcinoma (n=44), clear cell renal cell carcinoma (n=19), and papillary renal cell carcinoma (n=18)}}
\setlength{\tabcolsep}{0.2em}
{
\begin{tabular}{lrrrrrr}
\toprule
\multirow{2}{*}{\textbf{Organs}} & \multicolumn{3}{c}{\textbf{\# Slides}} & \multicolumn{3}{c}{\textbf{\# Patch Pairs}} \\ \cmidrule(lr){5-7} \cmidrule(lr){2-4} 
& \textbf{Train} & \textbf{Val} & \textbf{Test} &  \textbf{Train} & \textbf{Val} & \textbf{Test} \\ \midrule
Kidney$^\dagger$ & 48 & 15 & 18 & 125 & 41 & 41 \\ 
Head-neck & 13 & 5 & 6 & 27 & 9 & 10 \\ 
Prostate & 26 & 12 & 10 & 50 & 17 & 16 \\ 
Stomach & 15 & 6 & 5 & 36 & 12 & 12 \\ 
Endometrium & 38 & 13 & 13 & 86 & 29 & 25 \\ 
Bladder & 35 & 14 & 14 & 82 & 29 & 26 \\ \midrule
\textbf{Total} & 175 & 65 & 66 & 406 & 137 & 130 \\ \bottomrule
\end{tabular}
}
\vskip -2mm
\label{tab:dataset_size}
\end{table}

This section provides a detailed description of the OCELOT challenge, beginning with an overview of its underlying motivation, followed by a description of the dataset, evaluation, and major milestones of the event.

\subsection{Goal of the challenge}
The OCELOT 2023 challenge was organized to evaluate the performance of cell detection algorithms, enhanced by larger contexts and tissue semantics, applied to H\&E stained histopathology images. The unique aspect of this challenge was the introduction of a new dataset, whose configuration and annotation protocol allow for an explicit exploration of cell-tissue relationships for better cell detection tasks. To that end, small and large FoV patches were collected and annotated with labels for cell detection and tissue segmentation tasks, respectively. Crucially, the small FoV patches overlap and are contained within the larger ones.

Recent research by \cite{ryu2023ocelot} highlighted the spatial distribution of different cell types in histopathological images while developing models for cell detection tasks. Their study observed a notable correlation between the presence of tumor cells and cancerous areas, as well as between lymphocyte cells and stromal regions within the cancer microenvironment. These practical observations indicate a strong relationship between individual cell types and the surrounding tissue structures, suggesting that the spatial context of cells within tissues may play a critical role in accurate cell detection. This is in line with other works that attempt to explore the hierarchical nature of biological entities within a WSI \cite{pati2022hierarchical,van2021hooknet}. Apart from the OCELOT dataset, the PanopTILs dataset~\citep{Amgad2022} also provides overlapping cell nuclei and semantic tissue segmentation annotations for the assessment of tumor-infiltrating lymphocytes. However, dataset of the OCELOT Challenge offers tissue annotations for a substantially larger area surrounding the cell annotation region. As demonstrated by \cite{ryu2023ocelot}, this expanded field of view enables more effective modeling of cell-tissue relationships by capturing broader contextual information beyond the immediate cell microenvironment.

Moreover, \cite{ryu2023ocelot} empirically showed that incorporating cell-tissue hierarchical semantics into cell detection models can enhance their performance. By leveraging the spatial relationships between cells and the surrounding tissue context, the model achieved more reliable and better detection outcomes. This approach underscores the importance of considering the microenvironment in which cells reside, rather than focusing solely on isolated cell characteristics and morphology.

Based on afore-mentioned insights, the OCELOT 2023 challenge seeks to further investigate the complex interplay between cells and tissue structures in cell detection tasks. The challenge provides a benchmark for evaluating different approaches that incorporate cell-tissue semantic relationships, with the goal of advancing detection methods and improving the accuracy of automated cell detection in histopathology images.

\subsection{Dataset}

\begin{figure}[t]
    \centering
    \hfill
        \includegraphics[width=\linewidth]{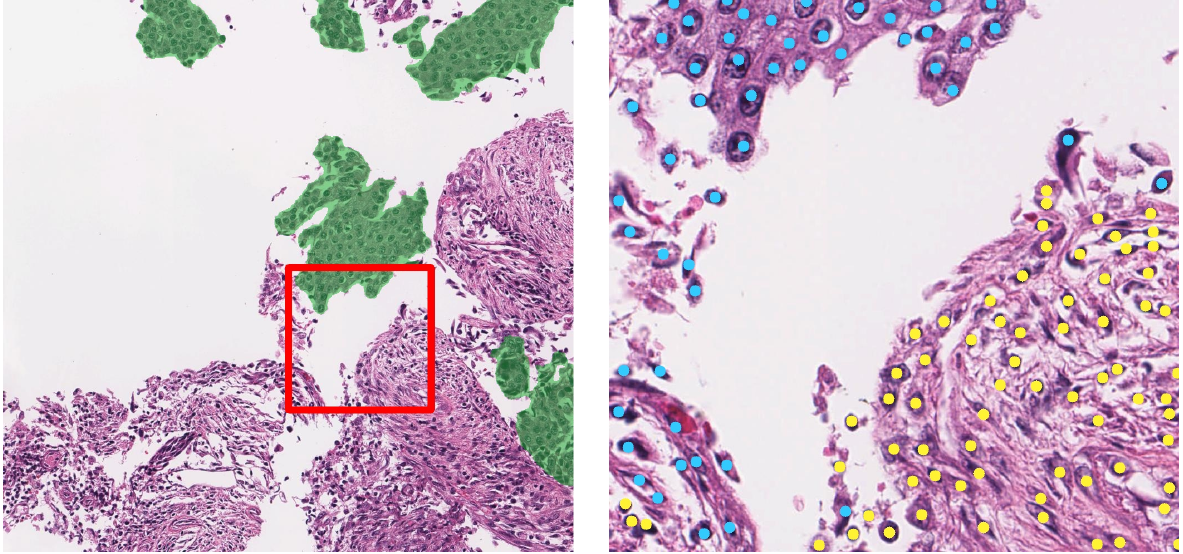}
    \hfill
    \vskip -7mm
    \caption{A sample from the \ours dataset includes two input patches with their respective annotations. The left side shows a large FoV patch with tissue segmentation (green indicating cancer), while the right side displays a small FoV patch with cell point annotations (blue dots for tumor cells, yellow for background cells). A red box outlines the small FoV patch’s position within the large FoV patch.}
    \label{figure:data_sample}
    \vspace{-0.3cm}
\end{figure}

\begin{figure*}[!ht]
  \centering
  \begin{minipage}[c]{0.27\textwidth}
    \centering
\includegraphics[width=\linewidth]{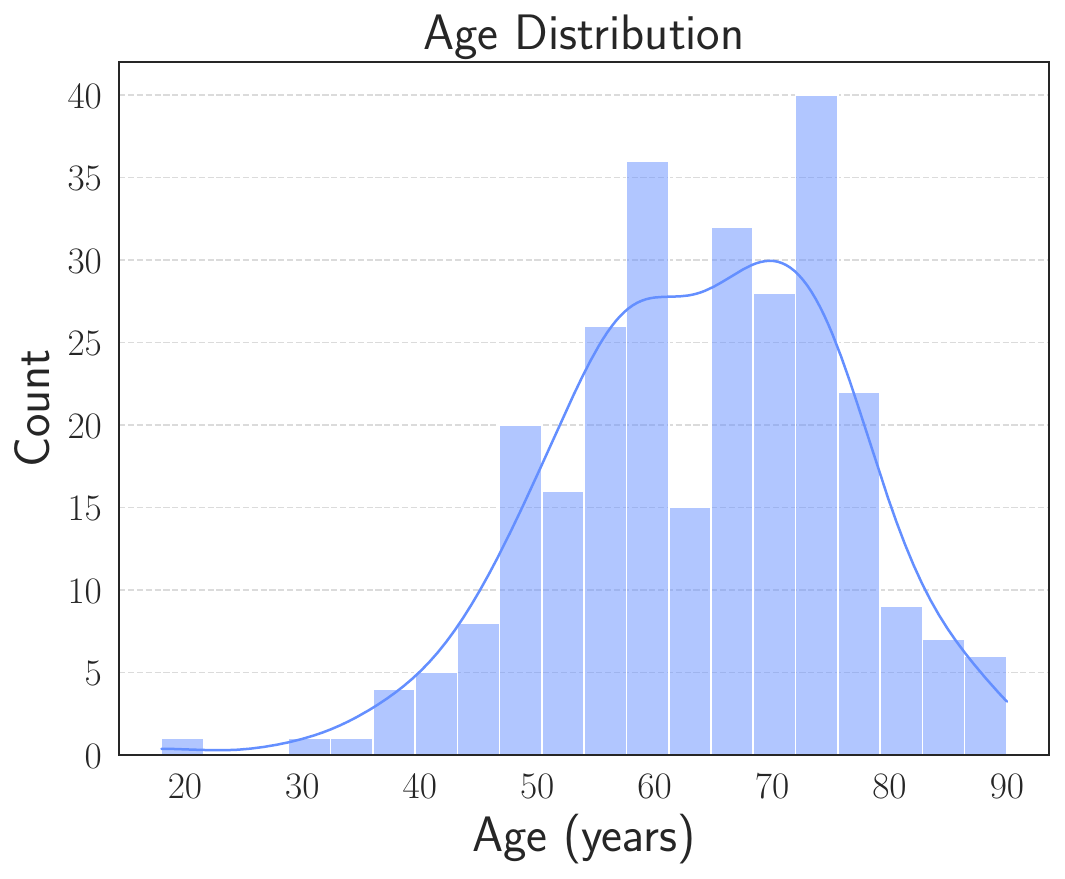}
\caption{Age distribution of challenge cohort.}
\label{fig:age_dist}
  \end{minipage}
  \hfill
  \begin{minipage}[c]{0.7\textwidth}
    \centering
    \includegraphics[width=\linewidth]{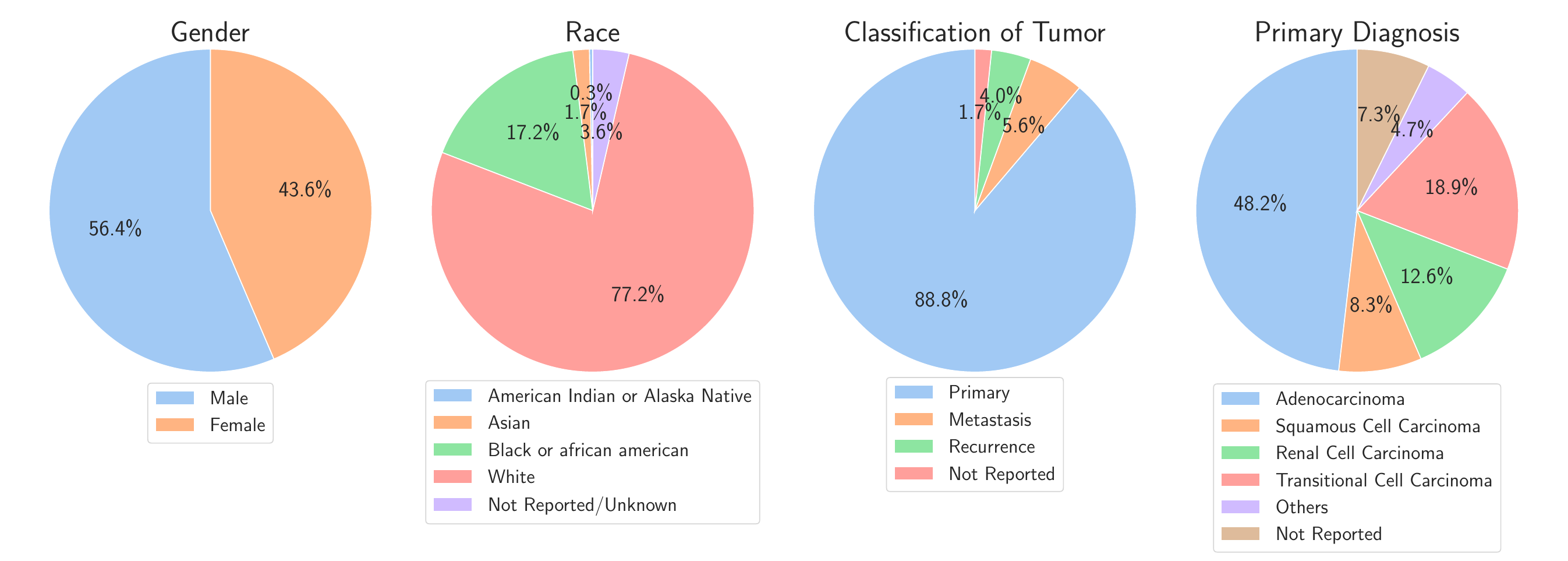}
    \caption{Population characteristics of the challenge cohort. From left to right, pie charts display the distribution of gender, race, classification of tumor and primary diagnosis. Some detailed categories have been grouped into larger categories. Metadata were obtained from the NIH-GDC portal (\url{https://portal.gdc.cancer.gov/}).}
    \label{fig:pies}
  \end{minipage}
\end{figure*}

The challenge dataset was acquired from 306 WSIs from TCGA, representing 6 distinct organs: kidney, head-and-neck, prostate, stomach, endometrium, and bladder. The specific number of slides for each organ is detailed in \autoref{tab:dataset_size}, and additional demographic and clinical characteristics of the challenge cohort are presented in  \cref{fig:age_dist} and \cref{fig:pies}. Each WSI was meticulously examined for selecting between one to three Regions of Interest (ROIs) for the tissue segmentation task. These ROIs were chosen based on their relevance and the quality of tissue representation. For the cell detection task, smaller ROIs were randomly selected from within these larger tissue ROIs, ensuring they were fully contained within the original selections. An example of this process is illustrated in Figure \ref{figure:data_sample}. As a result, the OCELOT dataset consists of 673 paired cell and tissue patches across the 6 organs, thus providing a robust basis for evaluating the interplay between tissue structures and cell detection. No additional patient information or image acquisition details were provided to the participants beyond the image data itself. Also, the raw data was provided to participants without any pre-processing to allow for diverse approaches in data preparation.

\subsubsection{Training, validation, and testing split}
The OCELOT dataset is divided into three subsets: training, validation, and testing, following a 6:2:2 ratio. Specifically, the training, validation, and test subsets consist of 400, 137, and 130 patch pairs, respectively. To prevent information leakage between subsets, the dataset is randomly split at the WSI level, ensuring that different patches from the same WSI are not included in more than one subset. Additionally, the 6:2:2 subset split ratio was maintained consistently across organs to ensure balanced data distribution. The distribution of WSIs and the corresponding number of patch pairs for each organ are detailed in \autoref{tab:dataset_size}.

\subsubsection{Patch configuration}
In cell detection tasks, models are designed to leverage fine-grained spatial information, which is crucial for accurately capturing detailed cell properties such as borders, shapes, colors, and opacity. In contrast, tissue segmentation necessitates a broader FoV to effectively comprehend the overall structural context. Consequently, we set the FoV sizes for cell detection and tissue segmentation at 1024 × 1024 pixels and 4096 × 4096 pixels, respectively, at a resolution of 0.2 Microns Per-Pixel (MPP). 
The tissue patches provided to participants were  down-sampled by a factor of 4 (1024 x 1024, 0.8 MPP)
\autoref{figure:data_sample} illustrates an example from the OCELOT dataset.

\begin{table}[t]
\small
\addtolength{\leftskip} {-2cm}
\addtolength{\rightskip}{-4cm}
\setlength{\tabcolsep}{0.2em}
\centering
\caption{Number of annotated pixels for the tissue segmentation task and number of annotated cells for the cell detection task in the OCELOT dataset.}
{
\begin{tabular}{lrrrclrrr}
\cmidrule[\heavyrulewidth]{1-4}\cmidrule[\heavyrulewidth]{6-9}
\multirow{2}[2]{*}{\textbf{}}  & \multicolumn{3}{c}{\textbf{\# Pixels}}                                        & \enspace & \multirow{2}[2]{*}{\textbf{}} & \multicolumn{3}{c}{\textbf{\# Cells}} \\ \cmidrule(l){2-4}\cmidrule(l){7-9}

& \multicolumn{1}{c}{\textbf{Train}}   & \multicolumn{1}{c}{\textbf{Val}} & \multicolumn{1}{c}{\textbf{Test}} &   &
   & \multicolumn{1}{c}{\textbf{Train}} & \multicolumn{1}{c}{\textbf{Val}} & \multicolumn{1}{c}{\textbf{Test}}\\ \cmidrule{1-4}\cmidrule{6-9}
\textit{BG}\enspace  & 237.4M  & 79.1M  & 71.6M  & &
\textit{TC} & 43.8K  & 16.3K & 12.9K \\
\textit{CA}  & 171.0M  & 57.8M & 58.8M &  & 
\textit{BC}  & 23.6K & 8.4K  & 9.7K  \\

\textit{UNK} & 17.3M & 6.7M & 6.1M &  & 
{} & {}  & {} & {} \\  
\cmidrule[\heavyrulewidth]{1-4}\cmidrule[\heavyrulewidth]{6-9}
\textbf{Total} & 425.7M & 143.6M & 136.3M &  & 
\textbf{Total} & 67.4K  & 24.7K & 22.6K \\  
\cmidrule[\heavyrulewidth]{1-4}\cmidrule[\heavyrulewidth]{6-9}
\multicolumn{4}{c}{(a) Tissue Annotations} & & \multicolumn{4}{c}{(b) Cell Annotations} \\
\end{tabular}
}
\label{tab:annotation_stats}
\end{table}

\subsubsection{Annotation}
For the cell detection patches, each cell is labeled as a point annotation, providing 2D coordinates and class labels. Cells are categorized into two classes: Tumor Cell (TC) and Background Cell (BC), where the latter represents all the non-tumor cells (e.g. lymphocytes, macrophages, or others). Over the entire dataset, 35.01\% and 64.99\% of cells were labeled as TC and BC, respectively. 
For the tissue patches, labels are provided as pixel-wise segmentation maps, where each pixel is labeled as either Cancer Area (CA) or Background (BG). In cases where the tissue class was ambiguous, pixels were labeled as Unknown (UNK). The distribution of these classes is 55.77\% BG, 40.17\% CA, and 4.06\% UNK. The number of cell counts and tissue pixels is denoted in \autoref{tab:annotation_stats}.

Annotations were made by 67 board-certified pathologists from various countries. For tissue patches, pathologists annotated pixel-wise tissue categories by marking cancer (CA) and non-cancer (BG) regions. Non-marked regions were automatically assigned as UNK. For cell patches, a \textit{three-pathologist consensus strategy} was employed to minimize the high inter-rater variability which is common in cell annotations. Initially, two pathologists independently annotated each cell patch. Subsequently, a third pathologist reviewed and reconciled the two sets of annotations, resolving any discrepancies to produce a final, unified annotation.

Potential sources of error in the annotations include inter-observer variability among pathologists and the inherent complexity of distinguishing cell types in densely packed regions. While we did not quantify these errors, they represent common challenges in computational pathology tasks.
\renewcommand{\algorithmicrequire}{\textbf{Input:}}
\renewcommand{\algorithmicensure}{\textbf{Output:}}
\begin{algorithm}[t]
\caption{Counting of True Positive (TP), False Positive (FP), and False Negative (FN) for one cell class}
\label{algo:tp_fp_fn_determination}
\begin{algorithmic}[1]
\Require Cell predictions $\mathcal{P}$, Ground-truth cells $\mathcal{G}$
\Ensure TP, FP, FN counts
\State $TP \gets 0$, $FP \gets 0$, $FN \gets 0$ 
\State Sort $\mathcal{P}$ by confidence score in descending order

\For{each prediction $p \in \mathcal{P}$}
    \State $g \gets \text{nearest\_gt}(p, \mathcal{G})$
    \If{$\text{dist}(p, g) > 3 \mu m$}
        \State $FP \gets FP + 1$
    \Else
        \State $TP \gets TP + 1$
        \State $\mathcal{G} \gets \mathcal{G} \setminus \{g\}$ \Comment{Exclude from $\mathcal{G}$ to prevent double-matching}
    \EndIf
\EndFor
\State $FN \gets |\mathcal{G}|$ \Comment{Remaining unmatched ground-truth cells are False Negatives}
\State\Return TP, FP, FN
\end{algorithmic}
\end{algorithm}
\subsection{Evaluation}
The mean F1 (mF1) score was employed to evaluate cell detection performance and establish rankings, representing the average F1 score across the two cell classes. 
This metric was selected for its balanced consideration of precision and recall, both critical in cell detection tasks. 
For each cell class, the F1 score is calculated as the harmonic mean of precision and recall, defined as $F1 = 2 \cdot (precision \cdot recall)/(precision + recall)$. 
Precision and recall are computed based on True Positive (TP), False Positive (FP), and False Negative (FN) detections, determined using a hit criterion of a 3 $\mu$m radius circle (15 pixels at 0.2 MPP) centered on each predicted cell. A prediction is considered a TP if a ground truth cell falls within this circle; otherwise, it is an FP. Unmatched ground truth cells are counted as FN. 
In cases of missing results for a test case, it is treated as having no prediction.
\autoref{algo:tp_fp_fn_determination} details the matching process between predicted and ground-truth cells for computing these quantities.
The organizers have provided the evaluation code\footnote{\url{https://github.com/lunit-io/ocelot23algo/tree/main/evaluation}} that implements the above process, allowing participants to assess their algorithms' performance before submission.

\subsection{Challenge policy and timeline}
The OCELOT challenge, sponsored by Lunit, was a satellite event at MICCAI 2023 organized through the Grand-Challenge platform\footnote{\url{https://ocelot2023.grand-challenge.org/}}.
Lunit provided funding for data annotation, prize allocation, and overall competition management. The challenge was a one-time event with a fixed submission deadline, structured into three distinct phases:

\begin{enumerate}
  \item \textbf{Training Phase (April 14 - June 4, 2023)}: during this initial phase, only the training set was open for download for algorithm development. No additional data was allowed for training and validation. Publicly available pre-trained models (e.g. ImageNet) were allowed under the condition that participants disclose their usage.
  
  \item \textbf{Validation Phase (June 5 - July 28, 2023)}: In this phase, participants evaluated their algorithms on a concealed validation set via the Grand-Challenge platform. Validation scores were made publicly accessible through the Leaderboard standings in Grand-Challenge. Each team was allowed to make up to 10 submissions during the validation phase.
  
  \item \textbf{Test Phase (July 31 - August 4, 2023)}: During the test phase, participants submitted their final version of the algorithm for evaluation against a hidden test set. The test scores remained confidential until their disclosure at the OCELOT 2023 challenge event at MICCAI 2023. Each team was allowed to make only 1 submission during the test phase. This sub-set was used for the final rankings.
\end{enumerate}

During the validation and test phases, participants were required to submit their algorithms as Docker containers, to ensure only automatic methods were employed. The final ranking was determined by the test set mF1 score, with prizes awarded to the top three teams.
To be eligible for prizes, participants were required to submit both their code and a short report at the test phase. These submissions are publicly available on the test leaderboard page\footnote{\url{https://ocelot2023.grand-challenge.org/evaluation/test/leaderboard/}} of the challenge website. 
The organizers were ineligible to participate in the challenge or receive prizes. While individuals from the same institutions as the organizers were allowed to participate, they were not eligible for prizes.

Following the test phase, the entire dataset, including all subsets, was released for research use only via the Zenodo platform\footnote{\url{https://zenodo.org/records/8417503}}.

Participants who submitted their solutions during the test phase were invited to submit a manuscript to be published under LNCS Springer. Each submission was reviewed by 2 organizers through the OpenReview\footnote{\url{https://openreview.net/group?id=MICCAI.org/2023/OCELOT}} platform. In the end, a total of 6 papers were published in \cite{OCELOTMICCAIProceedings} as proceedings of the MICCAI 2023 challenge. Up to three participants from the five top-performing teams were invited as co-authors of this manuscript.

\section{Submissions}
\label{sec:submissions}

\begin{table*}[t]
    \centering
    \newcommand{\cell}{\textbf{\textsf{C}}}
    \newcommand{\tissue}{\textbf{\textsf{T}}}
    \caption{An overview of the top performing methods submitted to the OCELOT challenge. If a given operation or technique is applied to only one of the cell or tissue models, it is indicated in parenthesis as \cell\xspace or \tissue, respectively; if not indicated, it was applied to both models. }
    \renewcommand{\arraystretch}{1.2}
    \resizebox{\textwidth}{!}{
        \begin{tabular}{llllllll}
            \toprule
            \multirow{2}{*}{\textbf{Category}}  & \multirow{2}{*}{\textbf{Method}}  & \multicolumn{6}{c}{\textbf{Team}} \\ 
            \hhline{~~------}
                &   & \saltfish             & \joshua               & \lysch                & \biototem                 & \yllab                & \cite{ryu2023ocelot}         \\ 
            \midrule
            
            \multirow{17}{*}{Data pre-processing}    
                & Input normalization                   
                    & Scale RGB intensity   & Macenko norm. (\tissue)    
                                                                    & \xmark                & \xmark                    & \xmark                & \xmark                \\ 
            \hhline{~-------}
                & \multirow{13}{*}{Data augmentation}    
                    & Resize                & Random crop           & Re-Scale              & Random rotation (360\degree, \tissue)
                                                                                                                        & Random Flip           & Random Flip           \\ 
                &   & Flip                  & Rotation              & Random crop           & Symmetric transformations (\tissue)
                                                                                                                        & Shift                 & Random rotation       \\
                &   & Rotation              & Scale                 & Random flip           & Gamma enhancements (\tissue)
                                                                                                                        & Scale                 & Gaussian blur         \\
                &   & Add Gaussian noise    & Horizontal flip       & Random rotation       & Contrast enhancements  (\tissue)
                                                                                                                        & Rotate                & Add Gaussian noise    \\
                &   & Gaussian smoothing    & Gaussian blur         & Brightness variation  & Histogram equalization (\tissue)
                                                                                                                        & Coarse dropout        &                       \\
                &   &                       & Add Gaussian noise    & Contrast variation    & Solarization  (\tissue)   & Add Gaussian noise    &                       \\
                &   &                       & Color jittering       &                       & Color augmentation in HSV  (\tissue)
                                                                                                                        &                       &                       \\
                &   &                       & Downscaling           &                       & Gaussian blur  (\tissue)  &                       &                       \\
                &   &                       &                       &                       & Random scale  (\tissue)   &                       &                       \\
                &   &                       &                       &                       & Random flip (\cell)       &                       &                       \\
                &   &                       &                       &                       & Add Gaussian noise (\cell)&                       &                       \\
                &   &                       &                       &                       & Brightness enhancements (\cell)
                                                                                                                        &                       &                       \\
                &   &                       &                       &                       & Random sharpening (\cell)&                       &                       \\
            \hhline{~-------}
                & \multirow{2}{*}{Cell label generation}    
                    & Draw disks            & Generate a heatmap    & Soft instance segm.   & Repel coding scheme
                                                                                                                        & Draw disks            & Draw disks            \\ [-3pt]
                &   &                       & w/ Gaussian           & w/ NuClick            &                           &                       &                       \\ 
            \hhline{~-------}
                & Dataset filtering 
                    & \xmark                & Exclude 43 samples (\cell)
                                                                    & \xmark                & Exclude 41 samples        & \xmark                & \xmark                \\ 
            \midrule
            
            \multirow{5}{*}{\makecell[l]{Modeling of cell-\\tissue relationship}}
                & \multirow{3}{*}{During training} 
                    & Elm.-wise add         & Concatenate           & Concatenate           & \xmark                    & \xmark                & Concatenate \\ [-3pt]
                &   & tissue pred. to       & tissue pred. to       & tissue pred. to       &                           &                       & tissue pred. to\\ [-3pt]
                &   & cell inter. feat.     & cell input            & cell input            &                           &                       & cell input/inter. feat.     \\ [3pt]
            \hhline{~-------}
                & \multirow{2}{*}{After training}   
                    & \xmark                & \xmark                & \xmark                & Adaptive/\textbf{Extreme} weight fusing 
                                                                                                                        & Revise cell class with       
                                                                                                                                                & \xmark                \\[-3pt]
                &   &                       &                       &                       &                           & background tissue pred.  
                                                                                                                                                &                       \\
            \midrule
            \multirow{16}{*}{Training mechanisms}
                & \makecell[l]{Training of cell and \\ tissue model}
                    & Joint training        & Individual training   & Individual training   & Individual training       & Individual training   & Joint training        \\
            \hhline{~-------}
                & \makecell[l]{\# samples in \\train/validation split}
                    & 360/40                & \makecell[l]{364/36 (\tissue), 321/36 (\cell)}      & 320/80                & 287/72                    & 320/80                & 406/137               \\
            \hhline{~-------}
                & \multirow{2}{*}{Model Architecture}
                    & cellViT               & SegFormer-B0 (\tissue)& DeepLabv3+            & UperNet w/ VAN (\tissue)   & FC-HarDNet            & DeepLab v3+    \\
                &   &                       & SegFormer-B2 (\cell)  & w/ ResNet50           & SFCN-OPI (\cell)           &                       & w/ ResNet34 \\
            \hhline{~-------}
                & Pre-trained weight 
                    & SAM                   & ImageNet              & ImageNet              & ImageNet                   & N/A                   & ImageNet              \\
            \hhline{~-------}
                & \multirow{2}{*}{Loss}
                    & Dice + CE             & CE (\tissue)           & CE (\tissue)         & Dice + CE (\tissue)        & WBCE + WIOU             & Dice                  \\
                &   &                       & MSE (\cell)            & WMSE (\cell)         & WBCE (\cell)               &                       &                       \\
            \hhline{~-------}
                & \multirow{2}{*}{Input size}
                    & 512x512               & 512x512               & 896x896               & 512x512 (\tissue)          & 1024x1024             & 1024x1024             \\
                &   &                       &                       &                       & 64x64 (\cell)              &                       &                       \\
            \hhline{~-------}
                & \multirow{2}{*}{Total epoch}
                    & 50                    & 1500 (\tissue)         & 100 (\tissue)        & 40 (\tissue)               & 1000 (\tissue)         & 300                   \\
                &   &                       & 250 (\cell)            & 150 (\cell)          & 300 (\cell)                & 300 (\cell)            &                       \\
            \hhline{~-------}
                & K-fold validation
                    & \xmark                & \xmark                & 5-fold                & \xmark                    & 5-fold (\cell)         & \xmark                \\
            \hhline{~-------}
                & \multirow{3}{*}{Misc.}
                    & Finetune w/ LoRA      & Tiling w/ crop margin & Over-sampling         & Data sampling  (\tissue)  &                       &                       \\
                &   &                       &                       &                       & Re-Train with hard samples (\tissue)
                                                                                                                        &                       &                       \\
                &   &                       &                       &                       & Overlapped inference      &                       &                       \\
                &   &                       &                       &                       & NMS (\cell)               &                       &                       \\
                                                            
            \midrule
            
            \multirow{2}{*}{Post-processing}
                & Test Time Augmentation
                    & \xmark                & \xmark                & Rotation \& flip      & \xmark                    & Flip                  & \xmark                \\
            \hhline{~-------}
                & Ensemble                               
                    & \cmark                & \xmark                & \cmark                & \xmark                    & \cmark                & \xmark                \\
            \bottomrule
            \multicolumn{8}{l}{\textbf{Abbreviations}: norm. - normalization, inter. - intermediate, pred. - prediction, feat. - feature, CE - Cross Entropy, BCE - Binary Cross Entropy, IOU - Intersection Over Union, MSE - Mean Square Error,}\\
            \multicolumn{8}{l}{WMSE - Weighted Mean Square Error, WBCE: Weighted Binary Cross Entropy, WIOU - Weighted Intersection Over Union, LS - Label Smoothing}\\
        \end{tabular}
    }
    \label{tab:approaches}
    \vspace{-4mm}
\end{table*}

During the challenge period from June 5\textsuperscript{th} to July 28\textsuperscript{th}, 2023, a total of 290 participants engaged in the competition.
Among them, 260 submissions were recorded during the validation phase, with an additional 23 submissions made during the test phase. 
This section presents a concise summary of the methodologies adopted by the five top-performing teams during the test phase: \saltfish (team saltfish), \joshua (individual participant), \lysch (individual participant), \biototem (team biototem), and \yllab (team yllab).\footnote{Here, we reference the manuscripts submitted to the challenge proceedings or the report submitted during the test phase. 
}
Additionally, we also include the approach proposed by the challenge organizers \cite{ryu2023ocelot}.
These methodologies predominantly revolve around leveraging the cell-tissue relationship in four main stages: data pre-processing, modeling of cell-tissue relationships, training mechanisms, and post-processing. 
\autoref{tab:approaches} comprehensively aggregates the key techniques of each approach.

\subsection{Key technical contributions}
Analysis of the top-performing submissions reveals several key technical components that contributed to their success.
First, innovative cell label generation approaches were employed, with the top three teams implementing different strategies: \cite{li2023enhancing} used fixed-radius disks, \cite{millward2023dense} employed Gaussian distributions to convey boundary uncertainty, and \cite{schoenpflug2023softctm} utilized Nuclick \cite{koohbanani2020nuclick} for precise nucleus boundary detection, demonstrating its superiority over disk representations in their ablation study.
Second, the three top-performing teams incorporated cell-tissue relationships during the training phase. Notably, approaches that integrated cell-tissue relationships during training consistently outperformed post-processing methods, achieving higher overall F1 scores, tumor cell-specific F1 scores, and precision metrics, while post-processing approaches demonstrated better recall at the expense of precision.
Third, training methodologies differed significantly, with only the first-place team \citep{li2023enhancing} implementing joint training of cell and tissue models, while others trained them separately. \cite{li2023enhancing} also uniquely leveraged the Segment Anything Model (SAM) \citep{kirillov2023segment} pre-trained weights with Low-Rank Adaptation (LoRA) \citep{hu2021lora} fine-tuning for efficient adaptation of this large foundation model.
Finally, model ensemble techniques were employed by three of the five top teams, including the top performer, indicating the importance of combining multiple model predictions to achieve optimal performance. 
The following subsections provide detailed descriptions of each team's approach.

\subsection{Data pre-processing}

WSIs are scanned by various scanners, leading to variations from one WSI to another, including color intensity, contrast, and sharpness. 
\saltfish addressed this by scaling the RGB intensity levels to the range of $[0.01, 0.99]$, while \joshua implemented Macenko normalization \citep{macenko2009method} to standardize the tissue patches.

Additionally, various data augmentation techniques were employed to enhance the generalization ability of the models.
Basic geometric augmentations such as 90-degree random rotation and random flips were commonly adopted. Some teams also integrated scale augmentations.
In the case of photometric augmentations, the addition of Gaussian noise, Gaussian smoothing (or blurring), and color jittering, were widely adopted. \biototem stood out for their extensive use of diverse augmentations, including a full range of rotation angles instead of limiting to 90-degree increments, solarization that inverts all pixel values above a threshold, and histogram equalization. Additionally, they employed separate models for processing cell and tissue inputs, applying distinct augmentation techniques to each of them.

In the challenge, cell detection data were annotated with the location and class of each cell. 
The prevailing approach adopted by most teams involved converting these cell labels into a segmentation map, framing the cell detection task as a segmentation problem, as advocated by \cite{SwiderskaChadaj2019LearningTD} and \cite{ryu2023ocelot}.
A common technique used for this conversion is drawing fixed-size disks at each cell's location and filling them with its class label.
\cite{ryu2023ocelot} and \saltfish utilized a fixed radius of 1.4 $\mu$m for this purpose. In contrast, some teams explored more accurate cell boundaries as segmentation labels. 
\joshua, for instance, employed Gaussian distributions instead of disks to convey uncertainty regarding the boundaries. 
They selected a standard deviation of approximately 1.14 $\mu$m, ensuring an 8 $\mu$m diameter within 7 standard deviations.
\lysch took a unique approach by inputting the provided point list into Nuclick~\citep{koohbanani2020nuclick} to obtain precise cell nucleus boundaries. 
Nuclick, a CNN-based approach, simplifies complex cell-level annotation by drawing cell nucleus boundaries around points clicked inside cells. 
Additionally, \biototem implemented the repel coding scheme proposed by \cite{liang2019enhanced}, which is designed to better distinguish neighboring cells in crowded regions by considering entropy and reversibility.

\subsection{Modeling of cell-tissue relationship}
The motivation for the challenge was \textit{modeling cell-tissue relationships}, with most teams reporting improved performance by incorporating it into their models. 
Two main strategies were used for leveraging these relationships: in-training and post-training.

\subsubsection{In-training approaches} 
\cite{ryu2023ocelot} proposed a tissue-prediction injection model as a method for leveraging cell-tissue relationships during training. 
This model integrates the predicted tissue probabilities into the cell detection branch through concatenation. 
Similarly, \saltfish experimented with various methods of combining cell and tissue information, exploring optimal integration points within the models. 
They ultimately chose to merge the prediction results from the tissue branch with intermediate feature maps extracted by the cell branch encoder using element-wise addition. 
\joshua and \lysch took a different approach by channel-wise concatenating the tissue predictions with the cell input images. 
Notably, the top three performing teams all utilized the cell-tissue relationship during training.

\subsubsection{Post-training approaches} 
Some teams refined the cell detection using tissue predictions through heuristic rules. 
\biototem proposed two alternatives: adaptive and extreme weight fusion. In both cases, the cell class probabilities are refined based on the original probabilities and the corresponding tissue segmentation probabilities. 
Specifically, the adaptive weight fusion method places greater reliance on cell predictions when the cell detection model is confident (i.e. when one class shows a high predicted probability).
Conversely, it relies more on tissue segmentation results when the cell detection model is less confident (i.e. when two classes have similar predicted probabilities). 
In contrast, the extreme weight fusion method strictly follows the corresponding tissue segmentation results by converting all cells within a predicted tissue region to a corresponding cell class.
Interestingly, their experiments showed that extreme weight fusion outperformed adaptive weight fusion. However, this approach has a limitation in that it categorizes all cells within CA as TC, an assumption that does not necessarily reflect the biological reality.

\subsection{Training mechanism}
The proposed methods were trained through various techniques.

First, teams employed different strategies for training their cell and tissue models. While \saltfish jointly trained both models, other teams trained the tissue model separately, using its output to either train or refine the cell model.
Jointly training the models can implicitly model the cell-tissue relationship, leading to better feature learning. 
However, it demands more training resources and task interference can destabilize optimization, necessitating careful loss tuning. 
Conversely, independently training the models simplifies the process by not requiring paired cell and tissue data, and reduces potential interference. 
However, this approach may not fully benefit from the interaction between cell and tissue information.

Additionally, various architectures were used. 
\saltfish employed variants of the cell detection-specialized transformer-based model~\citep{vaswani2017attention}, CellViT~\citep{horst2024cellvit}. 
\joshua also used a transformer-based segmentation model, SegFormer~\citep{xie2021segformer}, with the B0 variant for tissue segmentation and the B2 variant for cell detection. 
\lysch opted for the CNN-based architecture DeepLab v3+~\citep{chen2018encoder} for both cell and tissue models, using a ResNet50~\citep{he2016deep} backbone, similar to \cite{ryu2023ocelot}. 
\biototem utilized UperNet~\citep{xiao2018unified} with a VAN~\citep{guo2023visual} backbone for the tissue model. 
For the cell model, the authors used SFCN-OPI~\citep{zhou2018sfcn} which is a model proposed for nuclei detection.
\yllab used FC-HarDNet~\citep{chao2019hardnet}, which is an efficient segmentation model.

To optimize each model, diverse losses, input sizes, and custom techniques were employed by various teams. 
\lysch and \yllab applied 5-fold cross-validation during training to make better use of limited data and ensure more robust performance estimates.
\saltfish used the ViT~\citep{dosovitskiy2020image} weights from SAM~\citep{kirillov2023segment} as initialization and applied LoRA~\citep{hu2021lora} to efficiently adapt the model. 
Performance degradation can occur at image boundaries due to a lack of neighboring context. 
To address this issue, \joshua introduced a tiling approach with a crop margin, accepting predictions only from the center of the cropped image. 
\lysch and \biototem tackled class imbalance by over-sampling data. 
\lysch balanced the presence of background and tumor cells, while \biototem balanced the number of pixels in each tissue class. 
Additionally, \biototem employed a two-stage training process for the tissue model. First, they train one model and use it to identify hard samples within the validation split, such as those with low dice scores (less than 0.8). Secondly, the authors fine-tune the model with the original training data along with these hard samples. 
Further, they divided the entire image into smaller, overlapping images for inference, and applied Non-Maximum-Suppression (NMS) to the cell model. 

\subsection{Post-processing data}
Finally, various strategies were employed to boost performance after training the models. 
\lysch applied test-time data augmentation using rotation and flip, while \yllab used flip augmentation.
Both teams then ensembled multiple predictions obtained through test-time augmentation and k-fold cross-validation to produce a final prediction.
\saltfish trained four Tissue-CellViTs with identical architecture but different initializations, employing a dual-model ensemble approach for prediction. 
The ensemble method averaged outputs from the tissue branches of all models, combined this with features from the cell branch encoder, and then averaged the cell branch outputs to produce the final prediction.

\section{Results and Discussion}

\begin{figure*}[!ht]
  \centering
  \begin{minipage}[c]{0.57\textwidth}
    \captionof{table}{Cell detection performance of the challenge submissions and the cell-only baseline in terms of F1 score (\%). The first column shows the mean across classes, followed by the score of each class. BC and TC stand for Background Cell and Tumor Cell, respectively. Values in brackets are the 95\% confidence intervals. In the table, the best performing score in each column is marked in bold, while the second best is underlined.}
    \label{tab:overall_f1}
    \resizebox{\linewidth}{!}{
        \begin{tabular}{lrrr}
            \toprule
            \\[-0.8em]
              \multicolumn{4}{c}{\large\textbf{F1-score (\%)}} \\
              \\[-0.8em]
            \textbf{Team}  & \makecell[c]{Mean}     & \makecell[c]{BC}  & \makecell[c]{TC} \\
            \midrule
            \saltfish & \textbf{72.44 \enspace[69.04, 75.23]} & 67.35 \enspace[62.39, 71.32] & \textbf{77.53 \enspace[74.33, 80.31]} \\
            \joshua & \underline{72.21 \enspace[69.13, 74.84]} & \underline{67.55 \enspace[62.98, 71.29]} & \underline{76.87 \enspace[73.85, 79.57]} \\
            \lysch & 71.73 \enspace[68.58, 74.59] & 67.34 \enspace[62.90, 71.22] & 76.11 \enspace[72.57, 79.18] \\
            \biototem & 71.18 \enspace[67.68, 74.24] & \textbf{67.73 \enspace[62.97, 71.77]} & 74.64 \enspace[70.80, 77.91] \\
            \yllab & 69.92 \enspace[66.17, 73.15] & 65.27 \enspace[59.89, 69.68] & 74.57 \enspace[70.85, 77.69] \\
            \midrule
            Cell-only baseline & 63.54 \enspace[59.88, 66.76] & 57.99 \enspace[52.69, 62.59] & 69.10 \enspace[65.26, 72.50] \\
            \toprule
            \\[-0.8em]
             \multicolumn{4}{c}{\large\textbf{Precision (\%)}} \\
             \\[-0.8em]
            \textbf{Team}  & \makecell[c]{Mean}     & \makecell[c]{BC}  & \makecell[c]{TC} \\
            \midrule
            \saltfish & \textbf{74.85 \enspace[71.34, 77.68]} & \textbf{75.20 \enspace[70.00, 79.54]} & 74.50 \enspace[69.28, 78.85] \\
            \joshua & 74.38 \enspace[70.99, 77.25] & \underline{73.08 \enspace[67.72, 77.46]} & \underline{75.67 \enspace[70.68, 79.82]} \\
            \lysch & \underline{74.56 \enspace[71.12, 77.74]} & 71.81 \enspace[65.17, 77.61] & \textbf{77.31 \enspace[73.68, 80.42]} \\
            \biototem & 70.67 \enspace[67.07, 73.95] & 66.89 \enspace[60.76, 72.21] & 74.46 \enspace[69.88, 78.58] \\
            \yllab & 68.15 \enspace[64.37, 71.52] & 66.33 \enspace[60.35, 71.58] & 69.97 \enspace[64.67, 74.56] \\
            \midrule
            Cell-only baseline & 62.06 \enspace[58.53, 65.31] & 61.18 \enspace[55.02, 66.58] & 62.94 \enspace[57.15, 68.20] \\
            \toprule
            \\[-0.8em]
            \multicolumn{4}{c}{\large\textbf{Recall (\%)}} \\
            \\[-0.8em]
            \textbf{Team}  & \makecell[c]{Mean}     & \makecell[c]{BC}  & \makecell[c]{TC} \\
            \midrule
            \saltfish & 70.90 \enspace[67.52, 74.10] & 60.98 \enspace[54.89, 66.43] & \textbf{80.81 \enspace[77.05, 84.60]} \\
            \joshua & 70.45 \enspace[67.42, 73.36] & 62.79 \enspace[57.13, 67.76] & 78.11 \enspace[74.63, 81.61] \\
            \lysch & 69.17 \enspace[65.69, 72.52] & 63.40 \enspace[57.92, 68.49] & 74.94 \enspace[69.19, 80.17] \\
            \biototem & \underline{71.71 \enspace[67.96, 75.19]} & \textbf{68.59 \enspace[62.28, 74.09]} & 74.82 \enspace[69.05, 79.84] \\
            \yllab & \textbf{72.02 \enspace[68.25, 75.52]} & \underline{64.24 \enspace[57.13, 70.54]} & \underline{79.80 \enspace[75.25, 83.93]} \\
            \midrule
            Cell-only baseline & 65.85 \enspace[62.20, 69.44] & 55.12 \enspace[48.39, 61.48] & 76.59 \enspace[73.01, 80.07] \\
            \bottomrule
        \end{tabular}
    }
  \end{minipage}
  \hfill
  \begin{minipage}[c]{0.42\textwidth}
    \includegraphics[width=\linewidth]{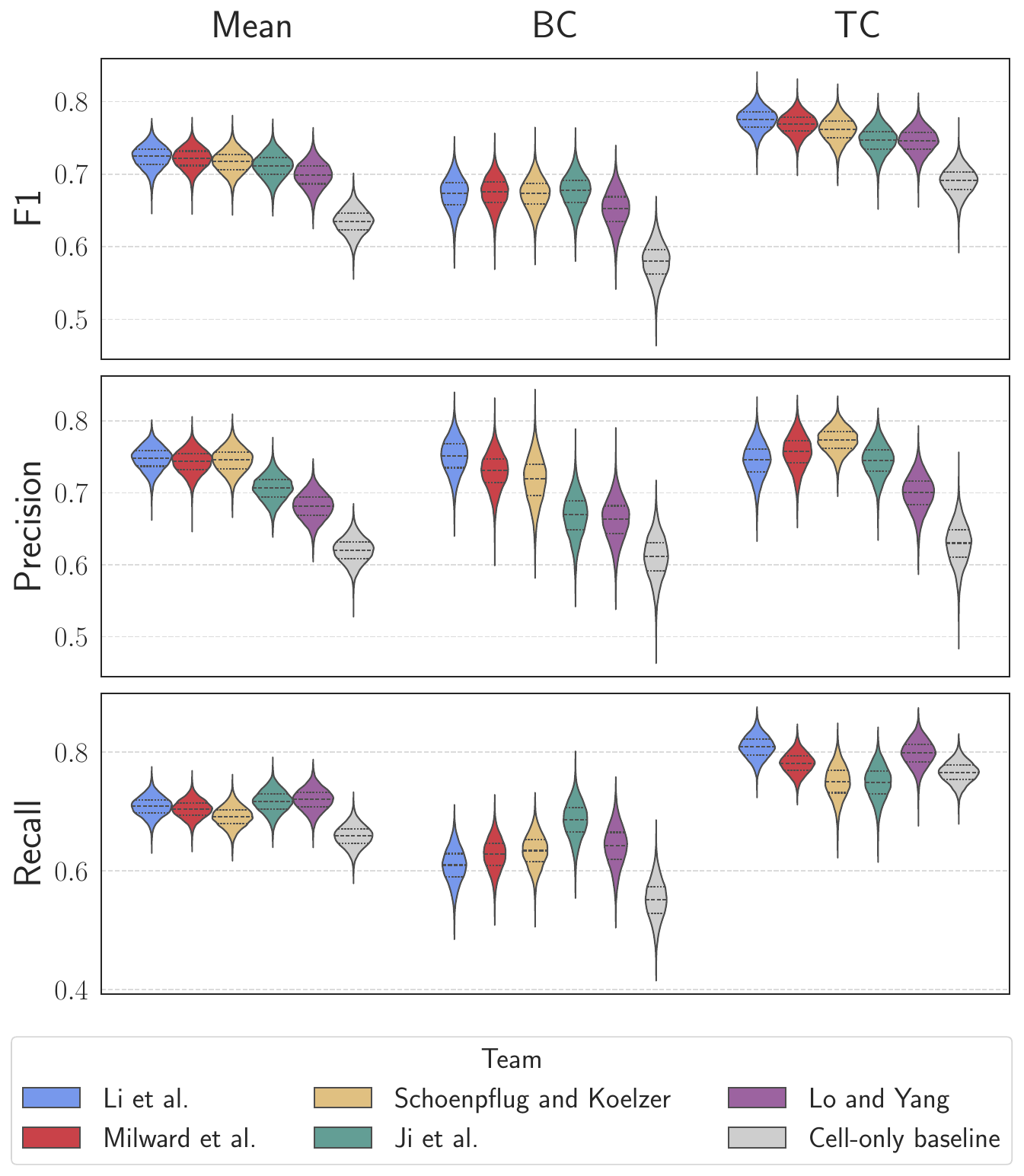}
    \caption{Violin plots representing the F1-scores, Precision, and Recall across the five top-performing teams and the cell-only baseline. Columns depict the mean, Background Cell (BC), and Tumor Cell (TC) results, respectively.}
    \label{fig:overall_perf}
  \end{minipage}
\end{figure*}
\label{sec:results}
This section presents the cell detection performance and analysis of the challenge submissions. 
We evaluate the results in the test set using multiple metrics: F1 score, precision, and recall. Furthermore, we present the results per individual cell class and inspect how the methods perform in the different tissue regions. All performance measures are reported with their 95\% confidence intervals computed through bootstrapping with 10,000 iterations.

\subsection{Overall results}

\begin{figure*}[!ht]
  \centering
  \begin{minipage}[c]{0.62\textwidth}
    \centering
\includegraphics[width=\linewidth]{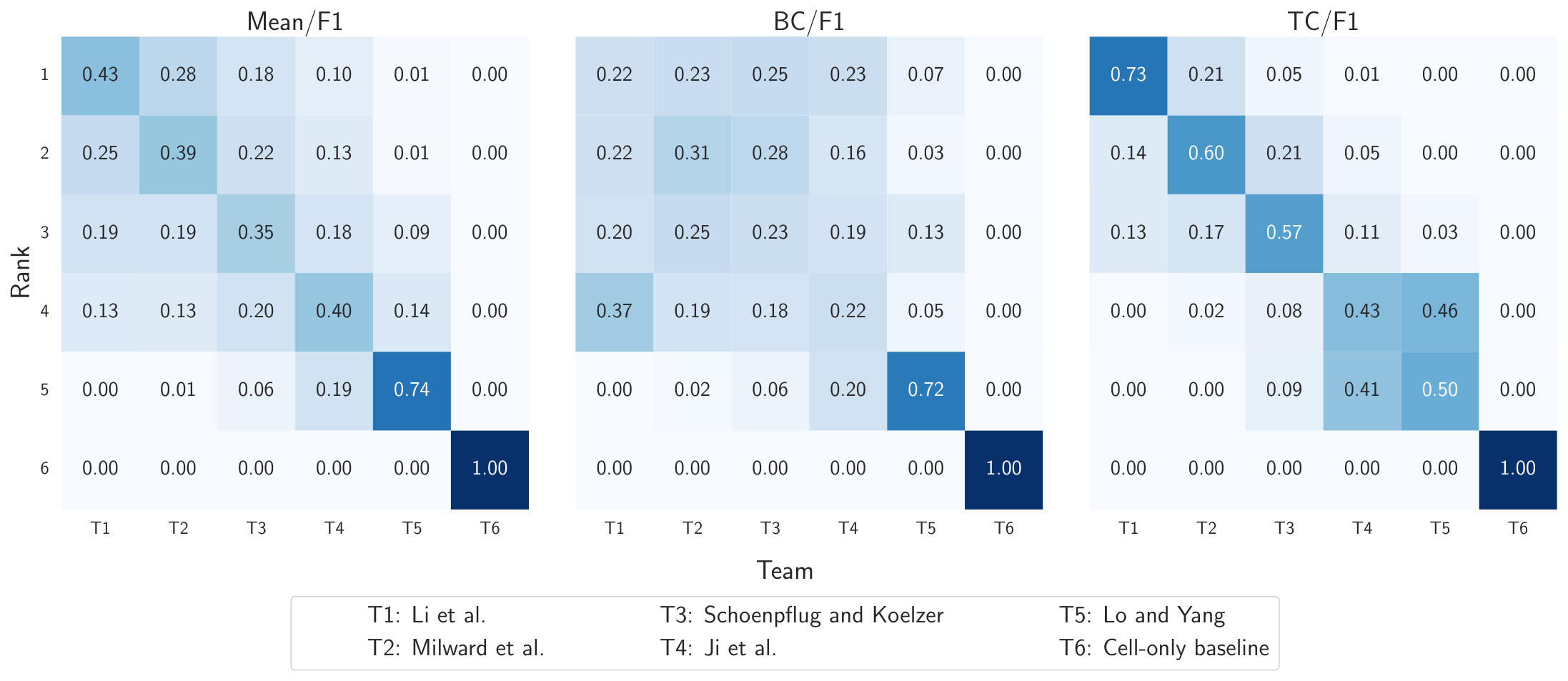}
\caption{Heatmaps showing the probability distribution of team rankings based on bootstrapping analysis. The panels display ranking probabilities for each team according to (left) mean F1 score across all cell classes, (middle) F1 score for Background Cells (BC), and (right) F1 score for Tumor Cells (TC). We observe the probability of each team (columns) being placed in each specific rank (rows).}
\label{fig:ranking_uncertainty}
  \end{minipage}
  \hfill
  \begin{minipage}[c]{0.35\textwidth}
    \centering
    \includegraphics[width=\linewidth]{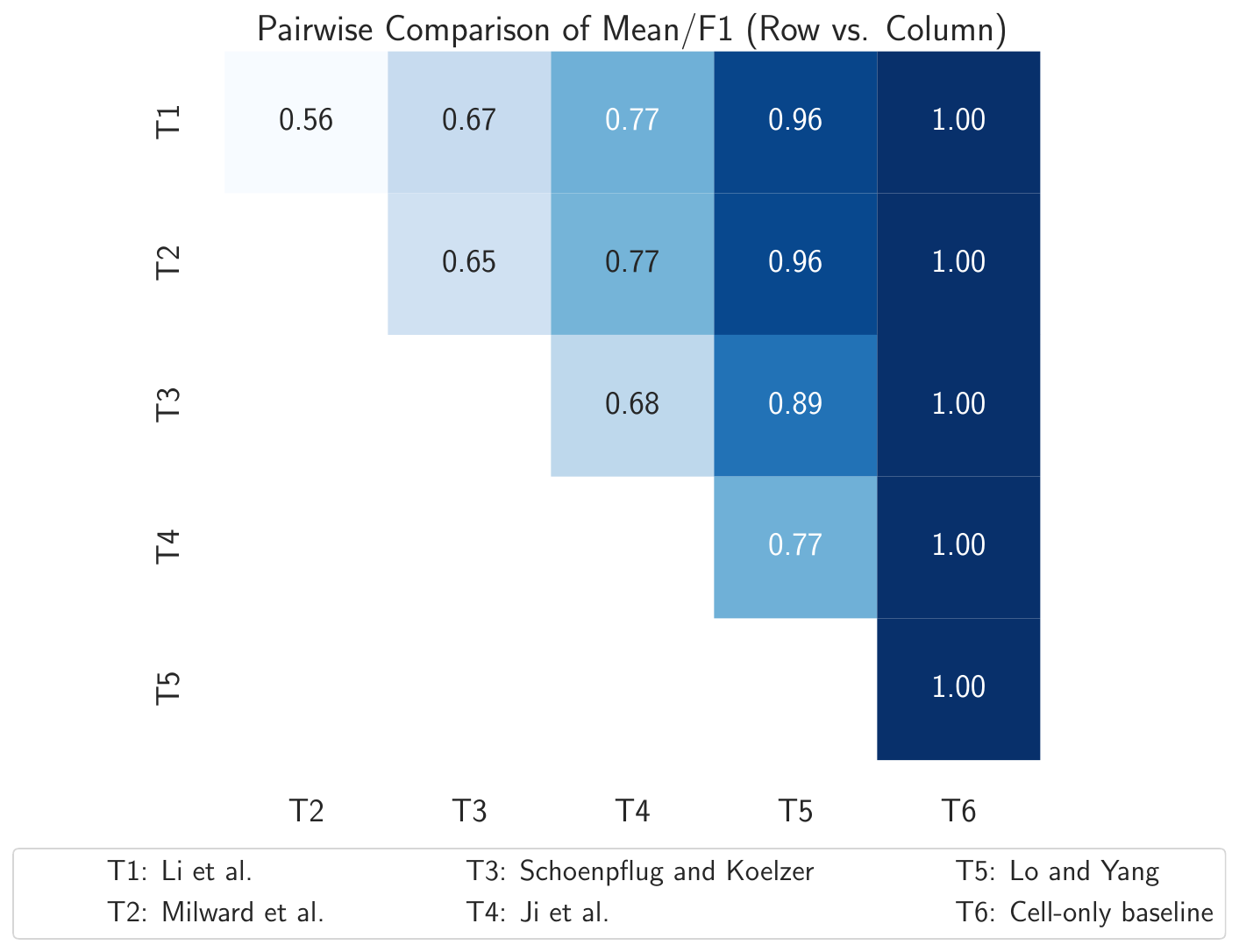}
    \caption{Heatmap displaying pairwise ranking probabilities, where each cell represents the probability that the row team outperforms the column team.}
    \label{fig:pairwise_comp}
  \end{minipage}
\end{figure*}

The cell detection performance of the five top-performing submissions is summarized in detail. \cref{tab:overall_f1} present the overall F1 scores, precision, and recall metrics. These tabular data are complemented by the corresponding violin plots in \autoref{fig:overall_perf}.  \UPD{Performance per primary site can be found in the supplementary material, Sec. \ref{sec: tissue-specific}.}
In addition, we also report a cell-only baseline, which serves as a control by not using cell-tissue relationships.

The cell-only baseline model is based on DeepLabV3+ architecture \cite{chen2017rethinking}, with a ResNet34 backbone \cite{he2016deep} pre-trained on ImageNet. During training, we used dice loss function \cite{milletari2016v} and the Adam optimizer \cite{KingmaB14} with an initial learning rate of 1e-4. The learning rate followed a step decay schedule, reducing by a factor of 0.3 in epochs 180 and 240, with training continuing for a total of 300 epochs. We applied dropout ($p=0.1$) and weight decay of 1e-5 to prevent overfitting.

All submitted solutions outperform the cell-only baseline across all metrics and cell types, with F1-score improvements ranging from 6 to 7.99 points. This further demonstrates the value of incorporating tissue context into cell detection and classification tasks and models. 
Additionally, these solutions consistently exhibit better F1 performance in both the TC and BC classes. 

Interestingly, the utilization of cell-tissue relationships yields a more substantial improvement in precision compared to recall (an average of 10.46 precision percentage points vs. 5.00 recall percentage points across all teams). Additionally, it enhances BC detection more than TC (an average 9.06 vs. 6.84 percentage points in F1 score). 
This differential improvement can be attributed to the strong association between cell types and tissue classes (TC with CA, BC with BG), which may provide a powerful prior for classification. We hypothesize that this prior leads to less False Positive detection of these cell classes, thus increasing precision.

We also observe that the confidence intervals for BC  are generally wider than for TC cells, suggesting more uncertainty in BC cell detection and classification. While TC represents tumor cells only, the BC class encompasses all other cells (i.e., non-TC cells), thus being more challenging due to its heterogeneity. The difficulty of the BC class is confirmed by the significantly lower F1 scores across all methods. This may suggest that there is further room for improvement within the BC class compared to the TC class.

Examining team performances reveals nuanced differences among the submissions. 
In \autoref{fig:overall_perf} we observe that the F1 scores of the top 3 teams are very close. Examining the cell classes, we conclude that the rankings were primarily determined by TC F1 scores since the F1 scores of BC were more similar across teams. 
A notable difference in results emerges between the top-3 and the 4\textsuperscript{th} and 5\textsuperscript{th} ranked teams. The top teams exhibit higher precision, while the remaining two teams achieve higher recall.
Among the top performers, \saltfish stands out with the highest BC precision of 75.20 and a balanced precision across cell classes. \joshua follows a similar pattern, albeit with slightly lower BC precision. \lysch distinguishes itself by achieving the highest TC precision. 
On the other hand, the 4\textsuperscript{th} and 5\textsuperscript{th} ranked teams showcase different strengths: \biototem leads in BC recall, while \yllab achieves the highest TC recall.

The way cell-tissue relationships are integrated within the proposed solutions appears to have an impact on the overall performance across teams. The top 3 teams incorporated those relationships during training, whereas the other two teams developed post-training heuristics. These results suggest that the models may learn how the different cells and tissues are related in more optimal ways.

We further assess the uncertainty in participant rankings. To do this, we use bootstrapping and recalculate the rankings based on each sample, thus estimating the probability of each team achieving a particular rank, as depicted in
\cref{fig:ranking_uncertainty}. 
In particular, when ranked according to the mean F1 score, all teams have the highest probability of maintaining their current rank. 
The cell-only baseline shows a very clear placement at the 6\textsuperscript{th} position. 
Furthermore, when examining TC F1 scores, we observe a distinct separation between the top three positions (1\textsuperscript{st}, 2\textsuperscript{nd}, and 3\textsuperscript{rd}), as well as a clear demarcation between the top-tier (ranks 1-3) and the second-tier (ranks 4-5) ranks. 
\cref{fig:pairwise_comp} displays the probability that a team in a given row achieves a better mean F1 score than a team in a given column, indicating that higher-ranked teams have higher probability of outperforming lower-ranked teams.

\subsection{Tissue region-specific cell detection performance}
\begin{figure*}[!ht]
  \centering
  \begin{minipage}[c]{0.42\textwidth}
    \centering
\includegraphics[width=\linewidth]{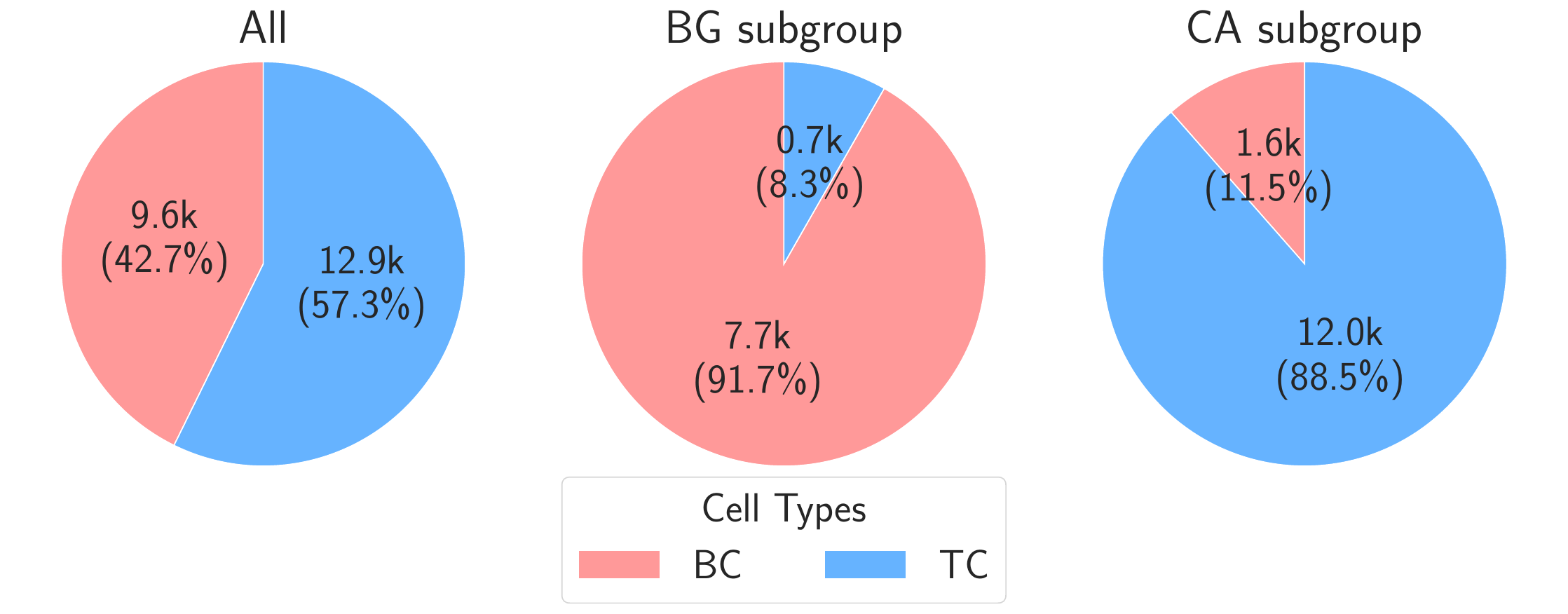}
\caption{Cell counts according to the tissue region class (subgroups) where the cells are located. The counts are based on the cell and tissue annotations. The cells located in Unknown (UNK) tissue regions are excluded. BC, TC, BG, and CA refer to Background Cell, Tumor Cell, Background, and Cancer Area, respectively.}
\label{fig:cell_count}
  \end{minipage}
  \hfill
  \begin{minipage}[c]{0.55\textwidth}
    \centering
    \includegraphics[width=\linewidth]{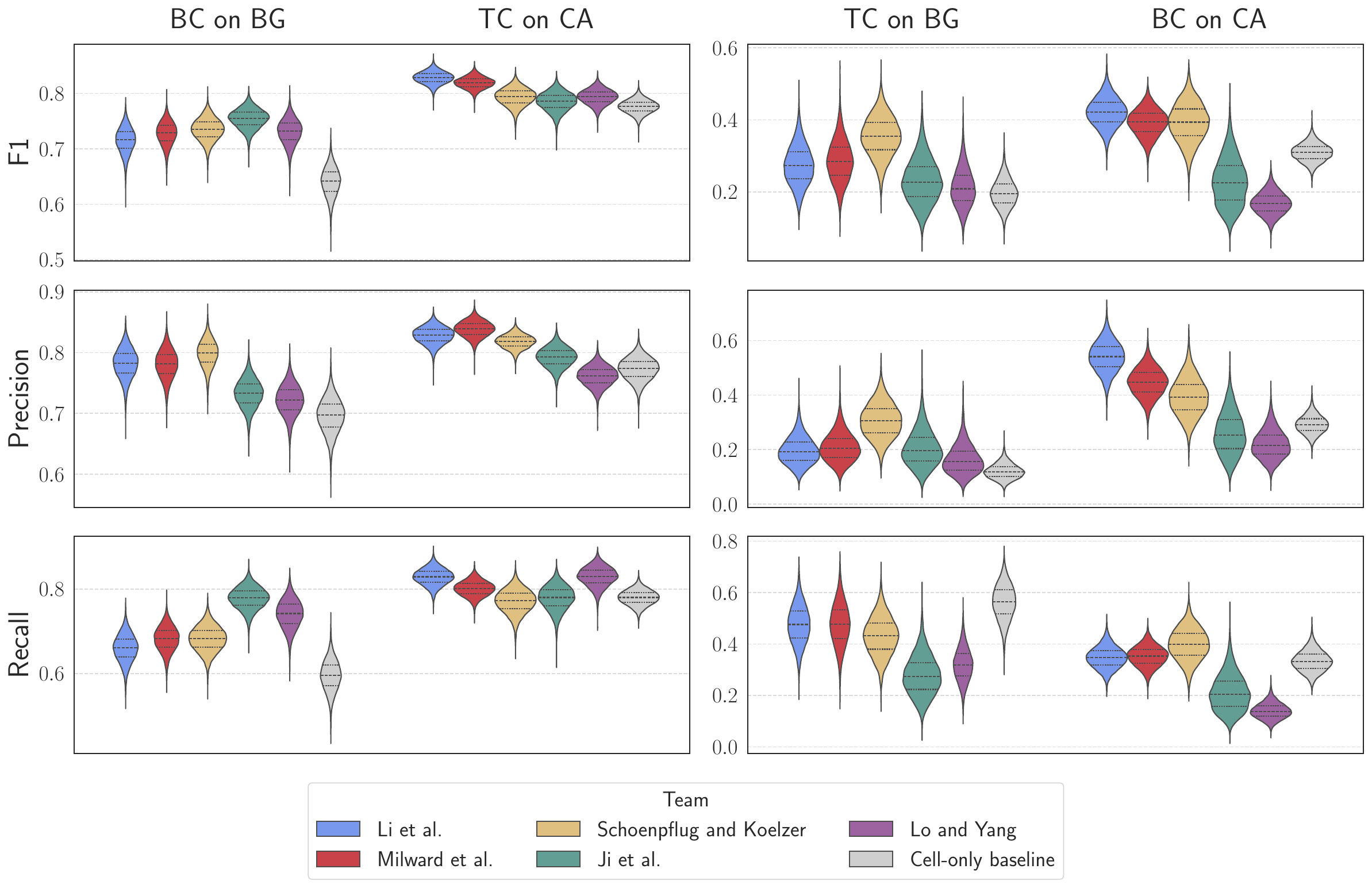}
    \caption{Violin plots showing per-team F1-scores, precision, and recall metrics for specific cell class predictions across annotated tissue regions. The left panels illustrate the common scenarios where Background Cells (BC) are located on Background regions (BG) and Tumor Cells (TC) are located on Cancer Areas (CA). The right panels represent the less common scenarios where Tumor Cells (TC) are found on Background regions (BG) and Background Cells (BC) appear in Cancer Areas (CA).}
    \label{fig:subgroup}
  \end{minipage}
\end{figure*}

To gain deeper insights into how cell-tissue relationships influence model predictions, we conduct a subgroup analysis based on the tissue region where the detected cells are located.
We categorize cells according to the tissue region class in their background (BG or CA) and evaluate performance for each subgroup separately.  
We use the tissue segmentation annotations to filter cells identified by the model predictions and cell annotations. In this way, for a given tissue region class, we examine the cells located in the same region. This analysis provides valuable insights into model performance in different tissue contexts and how the cell-tissue relationship influences cell detection. Importantly, this analysis was conducted using the tissue segmentation annotations from the Test sets, which were unavailable for the model training. 
In this way, this analysis serves as a proxy evaluation, providing insight into the behavior of the models.

The cell count distribution (\autoref{fig:cell_count}) reveals a strong correlation between cell types and tissue classes, but not an absolutely strict co-location. Indeed, 91.7\% of cells on BG tissue are BC and 88.5\% of cells on CA tissue are TC. The presence of TC on BG regions and BC on CA regions highlight the complexity of the cell-tissue relationship. Accurately detecting specific cell types in these less common cell-tissue pairings can be crucial in clinical contexts.
For instance, the identification of lymphocytes (a type of BC) within cancerous areas is critical, as tumor-infiltrating lymphocytes serve as a powerful biomarker in cancer prognosis and treatment response prediction \citep{park2022artificial,rakaee2023association}. This complex interplay between cells and their tissue context presents both challenges and opportunities for cell detection models. While strong correlations may aid in classification, models should be flexible enough to detect and correctly classify cells in atypical tissue contexts, which often carry significant clinical importance. This may also be one reason why methods that learn cell-tissue relationships in training achieved higher performance compared to more strict post-processing solutions.

\autoref{fig:subgroup} illustrate performances of each subgroup: all cells, cells detected in the BG tissue, and cells detected in the CA tissue. We observe that submissions generally outperform the cell-only baseline for BC detection in BG regions and TC detection in CA regions, as expected from the strong correlation observed in the cell count distribution. The performance gain is particularly pronounced in precision. For instance, \saltfish achieves a precision of 78.29\% for BC in BG, compared to 69.69\% of the baseline. 
For TC in CA, \joshua reaches 83.87\% precision, a substantial improvement over the baseline, at 77.27\%. 
This increase in precision suggests that the tissue context helps the models to make more confident and accurate positive predictions.

In contrast, the impact over recall is not as clear. While some teams show significant gains, others maintain similar or slightly lower recall compared to the baseline, especially for TC in CA. 
This indicates that leveraging tissue context primarily helps reducing false positive detection rather than increasing sensitivity.

Leveraging the cell-tissue relationship generally improves performance, but it may also introduce a side effect of degraded performance in less common cell-tissue pairings, specifically BC on CA and TC on BG. For instance, \biototem and \yllab achieve a strong performance for BC on BG regions (F1-scores of 75.54\% and 73.16\%, respectively), but obtain a decreased performance in relation to BC on CA regions (F1-scores dropping to 23.40\% and 16.87\%). 
This stark contrast suggests these models may have overfit to the dominant BC-BG relationship, potentially due to their heuristic modeling of cell-tissue relationships, potentially compromising their ability to detect BC in atypical contexts.

Moreover, all teams show degraded performance in TC on BG recall compared to the cell-only baseline. 
While the baseline achieves a recall of 56.30\% for TC on BG, the submissions range from 27.94\% (\biototem) to 47.71\% (\saltfish). 
This consistent drop in recall indicates that leveraging tissue context is beneficial overall, but can increase the risk of false negatives in atypical scenarios. 
Such a trade-off could have significant clinical implications, potentially missing crucial indicators like early-stage tumors or metastases.
\lysch stands out in this challenging scenario of TC on BG, achieving the highest F1-score (35.42\%), primarily due to superior precision (30.12\% compared to 11.93\%-20.44\% of the other teams). 
This suggests \lysch's approach may be more robust in handling atypical cell-tissue pairings, balancing the use of tissue context with cell-specific features.

Lastly, we observe higher variability in performance, evidenced by wider confidence intervals, for the minority class in each tissue type (TC in BG and BC in CA). 
This increased uncertainty reflects the inherent difficulty in consistently detecting and classifying cells in uncommon tissue contexts. These challenges may represent open questions for future research.

\section{Conclusion}
\label{sec:conclusion}

In clinical practice, pathologists need to understand the context and tissue architectures when identifying cells. While this is common practice, there was limited work around modeling cell-tissue relationships in computational pathology, in particular, due to the lack of annotated datasets for this purpose. The OCELOT 2023 challenge was organized to gather insights from the community and foster research around this problem.
Our analysis of the top-performing submissions reveals that incorporating tissue context consistently enhances overall performance across all metrics and cell types, validating the hypothesis proposed by \cite{ryu2023ocelot}.
Notably, the integration of cell-tissue relationships yielded more substantial improvements in precision compared to recall, suggesting that tissue context primarily aids in reducing false positive detection rather than increasing sensitivity.

The solutions generally exhibited better performance for Tumor Cells (TC) than Background Cells (BC), but they also demonstrated a trade-off in handling atypical cell-tissue pairings. 
This was observed in the marginal improvement, or even degraded performance, in detecting BC on Cancer Area (CA) and TC on Background (BG) tissues. 
This finding highlights the challenge of developing models that effectively balance the use of tissue context with cell-specific features, particularly in less common scenarios. However, such cases may hold significant clinical importance. Therefore, this is still an area for future research.

The challenge saw a diverse range of methodologies for leveraging cell-tissue relationships. 
While it is challenging to clearly attribute performance differences to specific methodological choices, we can gather some insights. 
For instance, teams that integrated cell-tissue relationships during the training process generally achieved higher precision and overall F1 scores. 
On the other hand, teams that applied heuristics during post-training showed better recall, but comparably lower F1 scores.

Future research could focus on developing more sophisticated methods for integrating cell and tissue information that maintain high performance in both typical and atypical cell-tissue pairings. 
In particular, improving model sensitivity to rare but clinically important cell-tissue combinations without compromising overall performance would be an important area for investigation. 
Additionally, refining the definitions of tissue contexts and cell classes may prove beneficial, potentially allowing models to leverage known biological priors more effectively. 
This approach could not only enhance the performance of the model, but also improve interpretability, which is crucial for clinical applications in computational pathology. Additionally, for this challenge we selected high-quality ROIs to validate the algorithmic advantages of modeling cell-tissue relationships. Hence, future work could further evaluate how the proposed methods perform in low-quality scenarios (e.g., ROIs with artifacts) that may exist in clinical practice, bringing these methods one step closer to real-world applications.

In conclusion, the OCELOT 2023 challenge has demonstrated the potential of leveraging cell-tissue relationships in computational pathology, while also highlighting important challenges that need to be addressed for reliable application in clinical settings. 
These insights will serve as a valuable foundation for future research and development in cell detection and classification in histopathology images.
\section*{Acknowledgments}

The results published here are in part based upon data generated by the TCGA Research Network: \url{https://www.cancer.gov/tcga}. 
We express our sincere gratitude to all participants of this challenge, with special thanks to those who submitted code and detailed reports during the test phase. 
Our appreciation goes to Lunit for their generous support, providing both operational funding and intellectual property in the form of data, which were instrumental to the success of this initiative. 
The authors also thank Mohammad Mostafavi and Seonwook Park for their contributions to the organization of the OCELOT 2023 Challenge.

\bibliography{refs}
\newpage

\appendix
\section*{Supplementary Material}
\setcounter{section}{0}
\setcounter{figure}{0}
\setcounter{table}{0}
\renewcommand{\thesection}{S\arabic{section}}
\renewcommand{\thefigure}{S\arabic{figure}}
\renewcommand{\thetable}{S\arabic{table}}

\section{Tissue-specific Performance}
\label{sec: tissue-specific}
\begin{figure*}[!ht]
  \centering
  \begin{minipage}[c]{0.8\textwidth}
    \includegraphics[width=\linewidth]{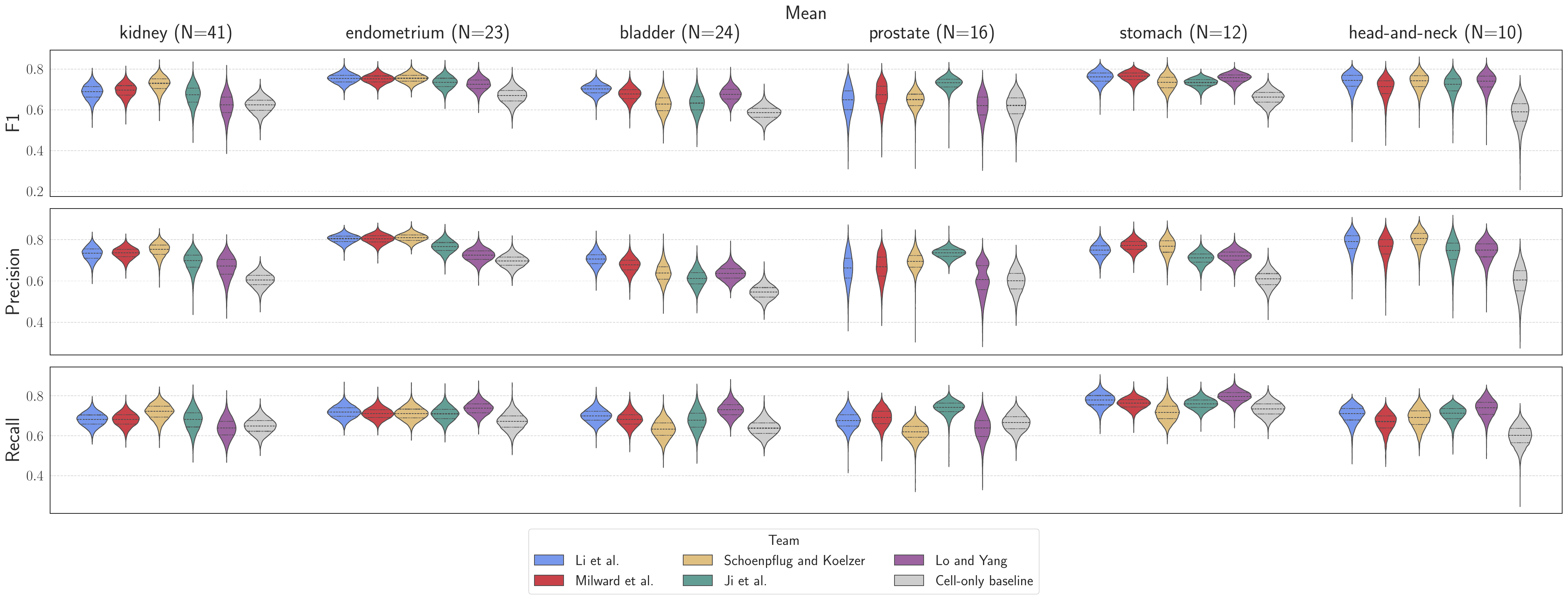}
    \caption{Tissue-specific performance by team (mean across all cell types) shown as violin plots. Each row displays the performance metrics (F1-score, precision, and recall). Each column represents a different tissue type, with the number of samples indicated for each tissue.}
    \label{fig:subgroup_organ_mean}
    \end{minipage}
\end{figure*}

\begin{figure*}[!ht]
  \centering
  \begin{minipage}[c]{0.8\textwidth}
    \includegraphics[width=\linewidth]{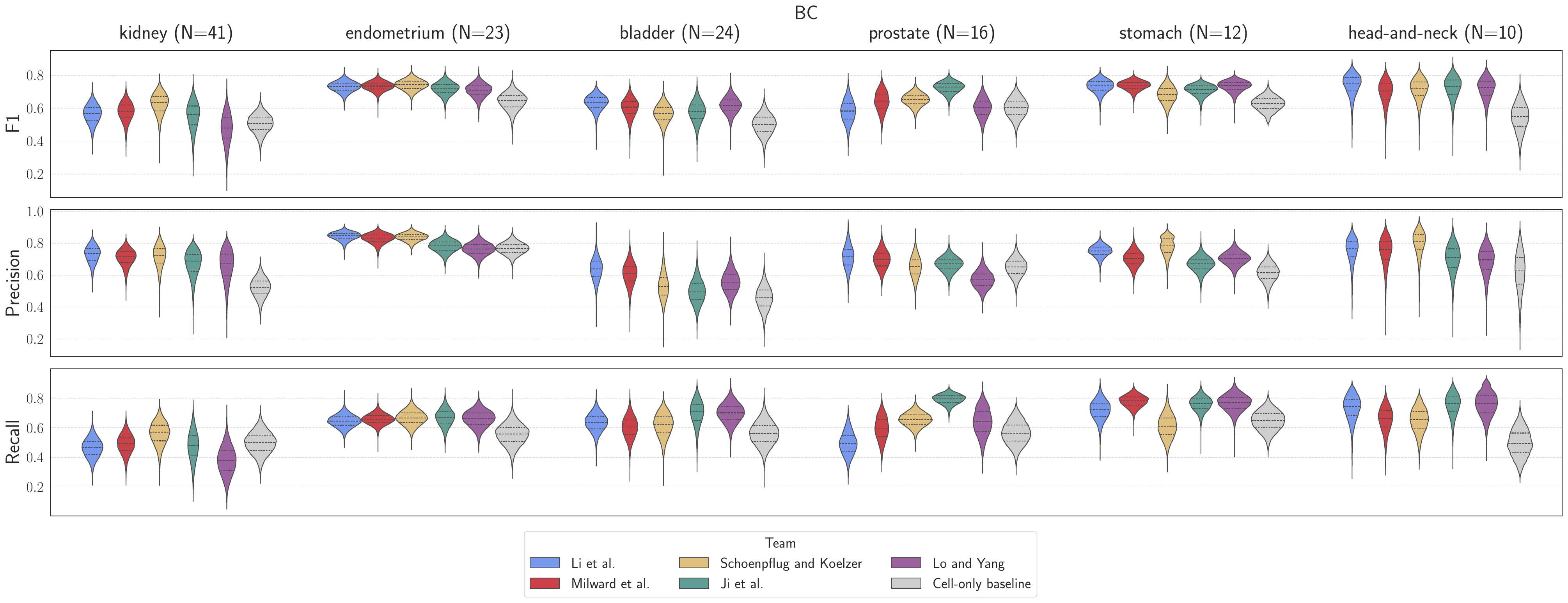}
    \caption{Tissue-specific performance of cell type Background Cell (BC) by team, shown as violin plots. Each row displays the performance metrics (F1-score, precision, and recall). Each column represents a different tissue type, with the number of samples indicated for each tissue.}
    \label{fig:subgroup_organ_bc}
    \end{minipage}
\end{figure*}

\begin{figure*}[!ht]
    \centering
    \begin{minipage}[c]{0.8\textwidth}
    \includegraphics[width=\linewidth]{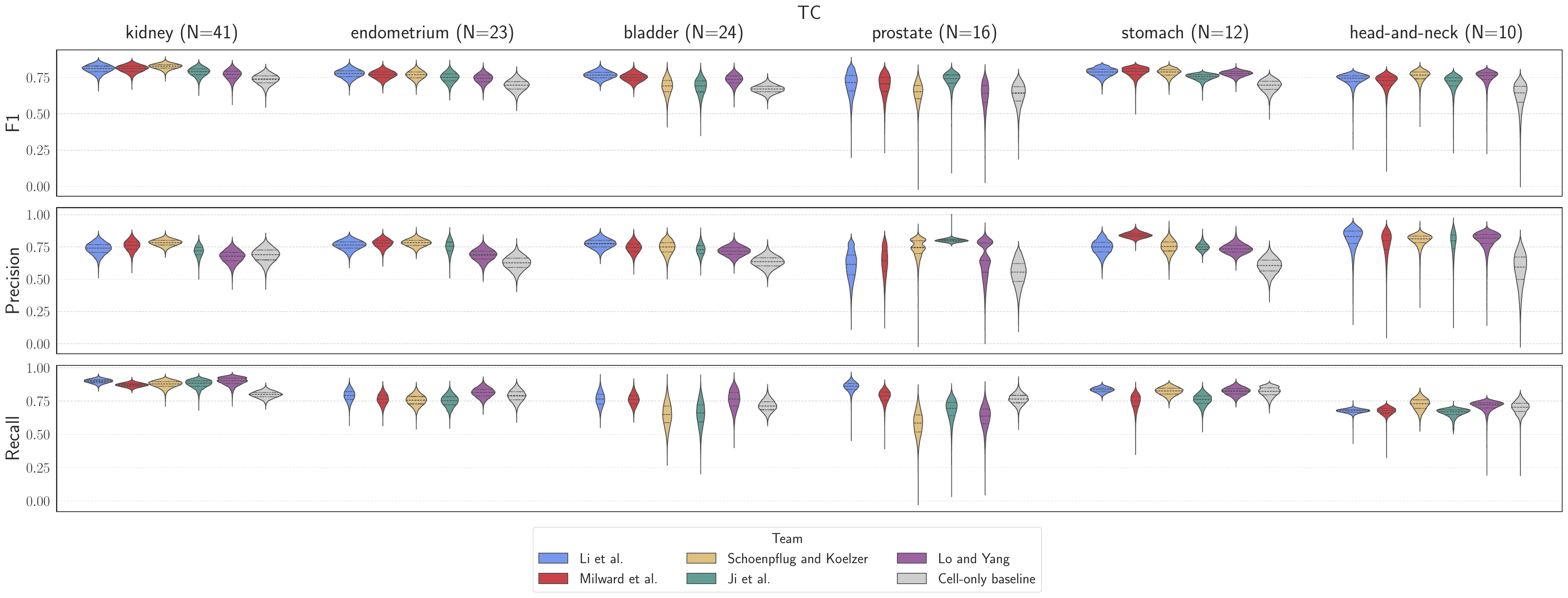}
    \caption{Tissue-specific performance of cell type Tumor Cell (TC) by team, shown as violin plots. Each row displays the performance metrics (F1-score, precision, and recall). Each column represents a different tissue type, with the number of samples indicated for each tissue.}
    \label{fig:subgroup_organ_tc}
    \end{minipage}
\end{figure*}

In this section, we present tissue-specific performance metrics across the six organ types (kidney, head-neck, prostate, stomach, endometrium, and bladder) in the OCELOT dataset. \cref{fig:subgroup_organ_mean}, \cref{fig:subgroup_organ_bc} and \cref{fig:subgroup_organ_tc} show the results of each team in terms of class average, background cell, and tumor cell, respectively. 

\section{Qualitative results}
\begin{figure*}[!ht]
  \centering
  \begin{minipage}[c]{0.8\textwidth}
    \includegraphics[width=\linewidth]{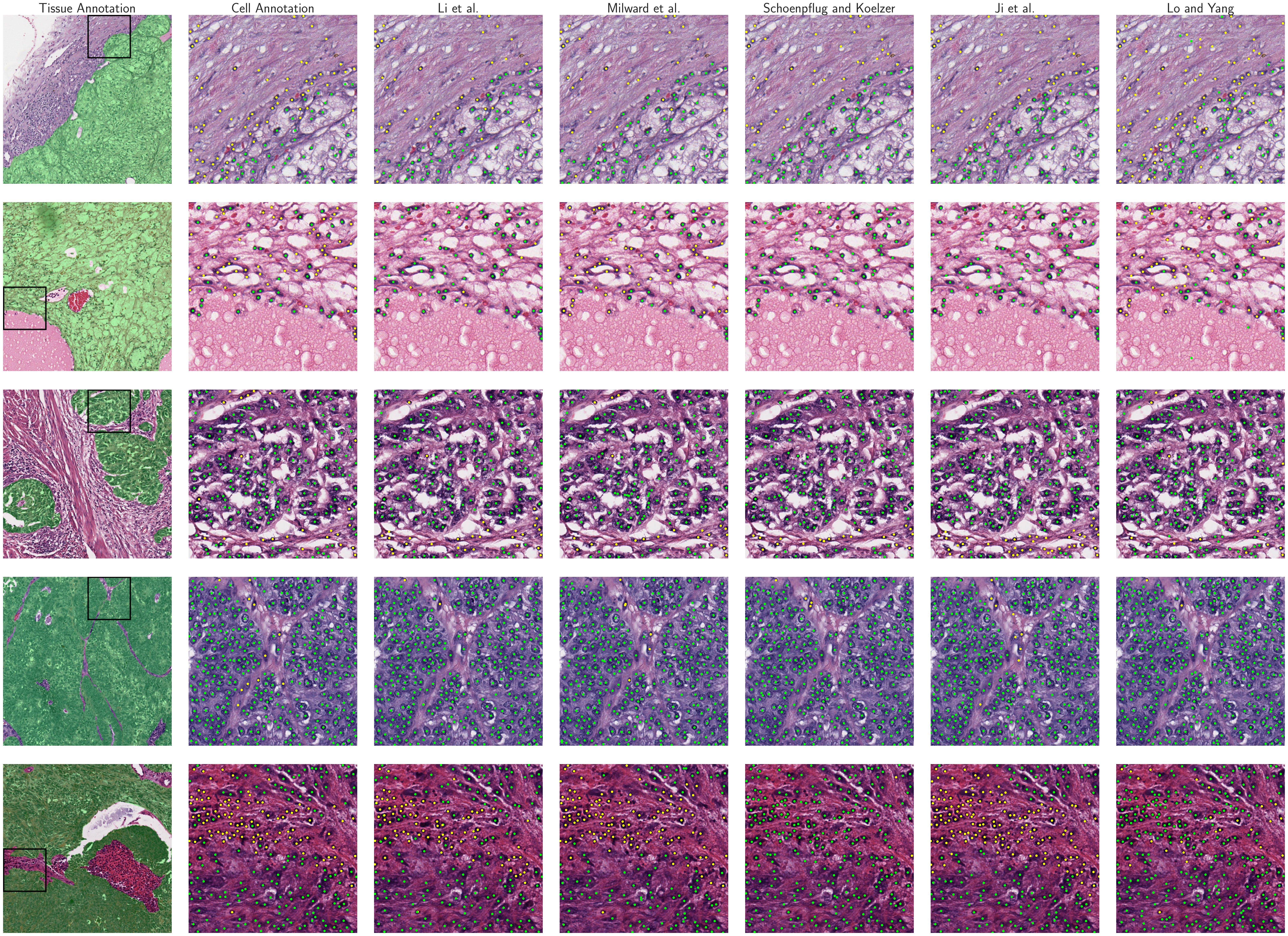}
    \caption{Visualization of annotations and participant submissions across five samples (indices 539, 542, 588, 591, and 612). The left two columns depict reference annotations: the first column shows Cancer Area (CA) highlighted in green with corresponding cell regions indicated by black boxes, while the second column displays Background Cells (BC) and Tumor Cells (TC) represented as yellow and green circles, respectively. The remaining columns present participant submissions using the same visual representation scheme as the reference cell annotations.}
    \label{fig:visualization}
    \end{minipage}
\end{figure*}

In this section, we visualize some of annotations and participant submissions in \cref{fig:visualization}. From the visualization, we can observe a few findings: First, tumor cells with small nucleus (second row) sometimes caused confusion in both the annotator and participant predictions. 
Second, in the third row, tumor cells at the bottom region were mistakenly identified as other cells without contextual awareness of the Cancer Area, precisely where annotator disagreements occurred. Notably, the models of the participants effectively leveraged surrounding context to correctly predict these lower-region tumor cells. 
Lastly, the fourth row demonstrates that solutions of the participants often classify other cells as tumor cells when they were interspersed among tumor cells.

\end{document}